# Clustering by connection center evolution

Xiurui Geng & Hairong Tang

Corresponding author: Xiurui Geng; Institute of Electronics, Chinese Academy of Sciences; E-mail: gengxr@sina.com.cn

**The determination of cluster centers generally depends on the scale that we use to analyze the data to be clustered. Inappropriate scale usually leads to unreasonable cluster centers and thus unreasonable results. In this study, we first consider the similarity of elements in the data as the connectivity of nodes in an undirected graph, then present the concept of a connection center and regard it as the cluster center of the data. Based on this definition, the determination of cluster centers and the assignment of class are very simple, natural and effective. One more crucial finding is that the cluster centers of different scales can be obtained easily by the different powers of a similarity matrix and the change of power from small to large leads to the dynamic evolution of cluster centers from local (microscopic) to global (microscopic). Further, in this process of evolution, the number of categories changes discontinuously, which means that the presented method can automatically skip the unreasonable number of clusters, suggest appropriate observation scales and provide corresponding cluster results.**

Determining cluster centers plays a critical role in cluster analysis. The K-means algorithm[1] iteratively calculates cluster centers by minimizing the within-cluster sum of squares. The distribution-based Gaussian mixture model[2] finds cluster centers through maximum likelihood estimation. The mean-shift algorithm[3,4] and the approach in ref.5 select the density peaks as cluster centers. Here we present a new and original viewpoint of cluster center determination. We consider that the cluster centers are evolving and dynamic rather than fixed and static, and whether a point is a cluster center depends on the scale we use to observe the data sets. For instance, the Sun can be considered as the center of the solar system, but it is obviously not the center of the galaxy system.

By mapping the data points to be clustered into the nodes of an undirected graph, we extend the concept of connectivity and employ it to dynamically determine the cluster centers. First of all, connectivity between nodes based on a pairwise similarity matrix is defined as follows.

The presented connectivity concept is informed by the definition of the number of walks[6]. Given a set of points $P = \{\mathbf{v}_1, \ldots, \mathbf{v}_n\}$ in $\mathbb{R}^L$. We use the element $s_{ij}$ of the pairwise similarity matrix $\mathbf{S}$ ($\mathbf{S} \in \mathbb{R}^{n \times n}$) to represent the *connectivity* between nodes $\mathbf{v}_i$ and $\mathbf{v}_j$, which can be considered as an extension of the number of walks with length 1 between $\mathbf{v}_i$ and $\mathbf{v}_j$ (see ref.7). Further, we use the entry $s_{ij}^{(k)}$ of the $k$-th power ($\mathbf{S}^k$) of the similarity matrix to define the *k-order connectivity* between $\mathbf{v}_i$ and $\mathbf{v}_j$ (denoted by $con^{(k)}(\mathbf{v}_i, \mathbf{v}_j)$), which can be approximately regarded as the number of walks with length $k$ between $\mathbf{v}_i$ and $\mathbf{v}_j$. Especially, the diagonal entry $s_{ii}^{(k)}$ is defined as *k-order connectivity* of node $\mathbf{v}_i$, which is denoted by $con^{(k)}(\mathbf{v}_i)$. For any node $\mathbf{v}_i$, we consider it to be a connection center of the graph and thus a cluster center of the data, if the number of walks with length $k$ between $\mathbf{v}_i$ and $\mathbf{v}_i$ is greater than that between $\mathbf{v}_i$ and $\mathbf{v}_j$. For instance, if each node can be considered as a station in a transportation network, it is natural that we use the connectivity of a node to represent how busy the station is. Obviously, the greater the connectivity, the more likely the station is to be a traffic center or a pivot. Specifically, the definition of cluster center can be strictly given by:

**Definition (cluster center)**: for one node $\mathbf{v}_i$, if it satisfies the following inequality, it will be a connection center of the graph and is defined as a *k-order cluster center*.

$$con^{(k)}(\mathbf{v}_i) \geq con^{(k)}(\mathbf{v}_i, \mathbf{v}_j), \ j = 1, \dots, n \tag{1}$$

For a pairwise similarity matrix $\mathbf{S}$ formed from a collection of nodes $\{\mathbf{v}_1, \dots, \mathbf{v}_n\}$, if a node satisfies inequality (1), we refer to it as a *diagonally maximal element*. If all the nodes satisfy inequality (1), $\mathbf{S}$ is named the *diagonally maximal matrix*.

After acquiring all the cluster centers, we next classify the remaining points to the corresponding cluster centers by the rules below. We assume that we have $m$ cluster centers $\mathbf{v}_{c_i}$, where $c_i \in \{1, 2, \cdots, n\}$ and $i = 1{:}m$. For any non-cluster-center point $\mathbf{v}_j$, in order to determine its assignment, we present the concept of the *k-order relative connectivity* between $\mathbf{v}_{c_i}$ and $\mathbf{v}_j$, defined as the ratio of $con^{(k)}(\mathbf{v}_{c_i}, \mathbf{v}_j)$ and $con^{(k)}(\mathbf{v}_{c_i})$ [i.e. $rcon^{(k)}(\mathbf{v}_{c_i}, \mathbf{v}_j) = con^{(k)}(\mathbf{v}_{c_i}, \mathbf{v}_j)/con^{(k)}(\mathbf{v}_{c_i})$]. The resulting classification rule is given by:

$$\mathbf{v}^* = \arg \max_{\mathbf{v}_{c_i}} \left( rcon^{(k)}(\mathbf{v}_{c_i}, \ \mathbf{v}_j) \right) \tag{2}$$

This means that $\mathbf{v}_j$ will be assigned to the cluster center ($\mathbf{v}^*$) that has the greatest *relative connectivity* with $\mathbf{v}_j$. This assignment is natural and reasonable, and makes the classification process simple and elegant.

In summary, for each $k$, cluster centers can be found by inequality (1), and then each of the remaining points is assigned to one cluster center according to equation (2). When $k = 1$, all points are cluster centers because the similarity matrix is generally a diagonal maximal matrix. Therefore, the initial number of categories is equivalent to the number of points, which is the extreme situation for a clustering problem. As $k$ increases, the acquired cluster centers and the corresponding cluster result gradually reflect the connectivity among the points from micro to macro. When $k$ approaches infinity, for a connected graph, the number of ultimate cluster centers will be generally reduced to one so that all data belong to the same cluster. Therefore, $k$ can be regarded as a scale-regulating factor which determines how we observe the data to be clustered, and the changing process of $k$ from small to large corresponds to the evolving process of connection center of the data from local to global. Therefore, we refer to this new method as connection center evolution (CCE).

Next, we use a simple 4-nodes example to illustrate the evolving process of connection centers as shown in Fig.1. First, a similarity matrix is constructed by a Gaussian kernel[3,4]. When $k = 1$, the number of connection centers is four and thus all the four points are cluster centers because the similarity matrix is a diagonally maximal matrix. Then it decreases to two with $\mathbf{v}_2$ and $\mathbf{v}_4$ being the local cluster centers when $k = 2$. Finally it is reduced to one when $k = 3$, and $\mathbf{v}_2$ is found to be the global cluster center. Interestingly, CCE cannot classify these points into 3 categories, which means that 3 is an unreasonable or prohibited number of clusters for CCE.

From the analysis mentioned above, given a collection of data points, CCE may provide various cluster results for different $k$. However, which one is reasonable? Next, we perform the CCE algorithm on the data points illustrated in Fig.2 and the corresponding number of clusters is recorded in Fig.2A. When $7 \leq k \leq 16$, the data points are grouped as 31 clusters (Fig.2A). In the next iteration interval ($25 \leq k \leq 95$), the data points are grouped as 6 clusters (Fig.2B). Finally, all the points are merged into one cluster (Fig.2C). Apparently, the results (31 and 6 clusters) are in line with people's intuition, so both of them are reasonable. In fact, the reasonable number of clusters can be suggested by the platforms of the curves as shown in Fig.2A.

Again, CCE jumps many states just as it does in the previous example. Especially, the number of clusters is reduced directly from 6 to 1 without any of the intermediate states (5, 4, 3, 2) . That is to say, CCE is able to intelligently knock out most unreasonable classification results.

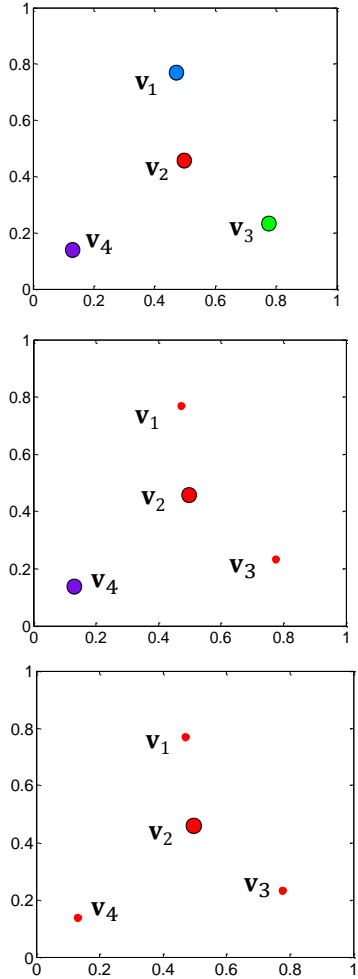

1. All the points are connection centers when $k = 1$

$$S = S^1 = \begin{bmatrix} 1.0000 & 0.7245 & 0.2852 & 0.1832 \\ 0.7245 & 1.0000 & 0.6547 & 0.4585 \\ 0.2852 & 0.6547 & 1.0000 & 0.2453 \\ 0.1832 & 0.4585 & 0.2453 & 1.0000 \end{bmatrix}$$

2. When $k = 2$, $\mathbf{v}_2$ and $\mathbf{v}_4$ become the connection centers because both $s_{22}^{(2)}$ and $s_{44}^{(2)}$ are the diagonally maximal elements of $\mathbf{S}^2$. $\mathbf{v}_1$ and $\mathbf{v}_3$ are both assigned to $\mathbf{v}_2$ according to (2), and $\mathbf{v}_4$ remains isolated.

$$S^2 = \begin{bmatrix} 1.6398 & 1.7197 & 1.0897 & 0.7686 \\ 1.7197 & 2.1637 & 1.6285 & 1.2103 \\ 1.0897 & 1.6285 & 1.5701 & 0.8430 \\ 0.7686 & 1.2103 & 0.8430 & 1.3039 \end{bmatrix}$$

3. When $k = 3$, the global cluster center ($\mathbf{v}_2$) is found because $s_{22}^{(3)}$ is the only diagonally maximal element of $\mathbf{S}^3$. The corresponding number of classifications is 1.

$$S^3 = \begin{bmatrix} 3.3372 & 3.9734 & 2.8718 & 2.1248 \\ 3.9734 & 5.0306 & 3.8324 & 2.9168 \\ 2.8718 & 3.8324 & 3.1539 & 2.1744 \\ 2.1248 & 2.9168 & 2.1744 & 2.2064 \end{bmatrix}$$

**Figure 1 | a simple example for CCE.** Here, a Gaussian kernel $s_{ij} = \exp(-\|\mathbf{v}_i - \mathbf{v}_j\|^2/\sigma^2), \sigma = 0.55$ is used. Connection centers are marked with bigger dots. Other points are marked with smaller ones and colored according to the cluster centers which they are assigned to.

For comparison, we apply k-means[1], mean-shift[3,4] and spectral cluster methods[8-10] to the data set mentioned above (Fig.S1-S4). A different number of clusters (6 and 31) should be specified to obtain different cluster results for the k-means and spectral cluster methods. Similarly, two different bandwidths are needed for means-shift in order to obtain two cluster results (6 clusters and 31 clusters) in different scales. From their cluster results, even though the correct parameters are provided, the results of 31 clusters for the k-means and spectral cluster methods are still unreasonable.

We also apply CCE to some widely used data sets[11]. The cluster results (see Fig. S5) demonstrate that CCE can be applicable to a variety of data with different shapes and distributions.

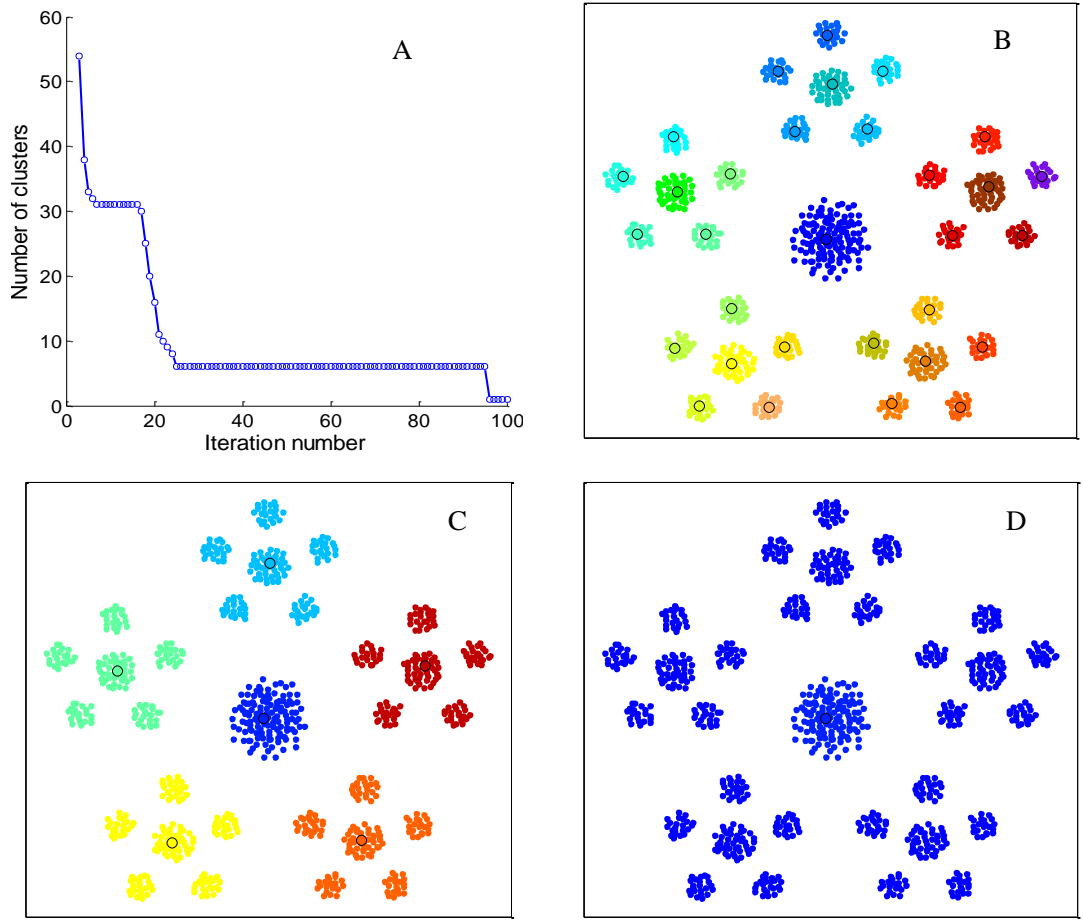

**Figure 2 | Result for a synthetic data set, using a Gaussian kernel.** When $k$ increases, the points are grouped as different clusters. Connection centers are marked with black circles and other points are colored according to the cluster centers which they are assigned to. (A) The number of clusters versus iteration number. (B) Iteration #7~16, 31 clusters. (C) Iteration #25~95, 6 clusters. (D) Iteration #96~, one cluster.

One interesting finding is that there exists a wonderful relationship between the spectral clustering algorithm and CCE, which is presented in the following theorem [see proof in ref.12].

**Theorem**: Given a collection of data points to be clustered, the square root of the diagonal vector of the $k$–th power of its similarity matrix is proportional to the eigenvector ($\mathbf{u}_1$) with a maximal eigenvalue of $\mathbf{S}$ when $k$ tends to positive infinity. i.e. $\mathrm{diag}(\sqrt{\mathbf{S}^k}) \propto \mathbf{u}_1, k \to \infty$.

Since CCE can find the global connection center of the data when $k$ is large enough, this theorem indicates that all the eigenvector-based clustering methods are inclined to focus on the global connectivity of a graph, which is not enough in many cases. Fortunately, the real data to be clustered usually corresponds to a multi-partite graph. When the similarity matrix is formed as a multi-block diagonal matrix, the connection center of each block can still be found based on the above theory. This may be the main reason why spectral clustering methods can succeed in many cases.

Next, we apply CCE to Olivette Face Database in which there are 10 different images of 40 distinct subjects[13]. Just like refs.1,14, the distance matrix is built by the method in ref.15. Here a Gaussian kernel with variance $\sigma = 0.05$ is used to construct the similarity matrix, and the matrix is normalized according to ref.9. The clusters with only one or two elements are judged as noise and thus are removed. From the Fig.3A and Fig.3B, we can see that, after about the 340th iteration, the number of clusters and several performance indices all tend to be stable and the corresponding results are encouraging. For instance, from the 481st to 510th iteration, CCE gets 128 clusters and 56 clusters before and after removing noise, respectively, among which not a single cluster encompasses images of two or more different subjects. Surprisingly, 8 subjects have been identified integrally (#2, 7, 11, 12, 14, 27, 30, 38) and 15 subjects have been recognized with no less than 7 images. In addition, 301 images have been assigned and only 3 in 40 subjects are not identified (#1, 13, 39). From the 511th to 530th iteration, the result is further improved. The 18th subject which had previously split into two clusters have merged so that 9 subjects have been identified integrally (#2, 7, 11, 12, 14, 18, 27, 30, 38). However, with the increase of iterations, two or more different subjects may be merged into one cluster. For instance, in the 531st iteration, the 10th image of the 10th subject is assigned to the integral class made of the 7th subject. As expected, all images will finally be merged into one cluster after enough iterations. The result of the 516th iteration is shown in Fig.3C.

From the above results, we find that CCE can work well on clustering the Olivette faces. In the following, we discuss the effect of the variance in Gaussian kernel on the performance of CCE. Obviously, how the variance is set determines the speed of convergence of CCE. The smaller one implies that the evolution of connection centers will proceed at a lower speed. It makes the number of clusters decrease more slowly and therefore possibly results in more long-lasting platforms, which usually correspond to the desired cluster results. On the other hand, the larger variance may cause the number of clusters to decrease steeply and thus greatly save time, but it may leap over some valuable intermediate results. For a compromise between performance and speed, a moderating value of 0.05 is suggested here.

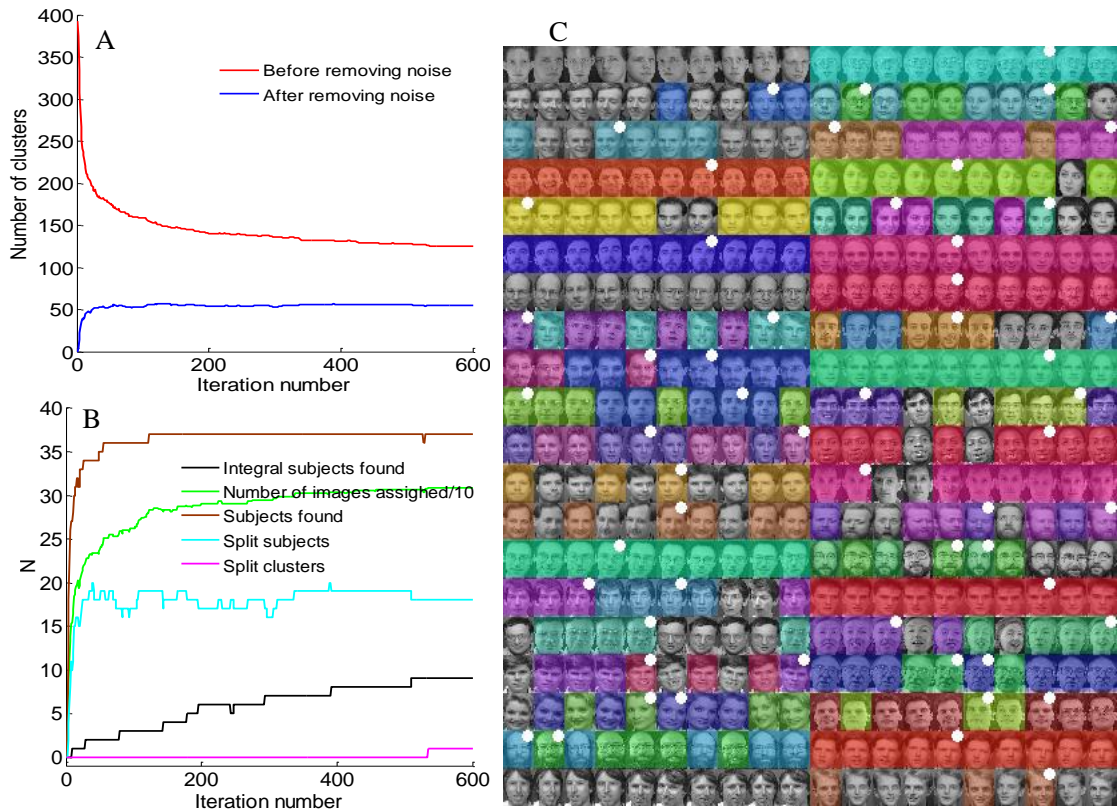

**Figure 3 | Clustering Olivetti faces.** (A) The number of clusters before (red) and after (blue) removing noises, versus iteration number. (B) The performance of CCE as a function of iterations for noise-removing data: the number of subjects identified integrally (black), the number of assigned images divided by 10 (green), the number of subjects identified as individuals (brown), the number of subjects split in more than one cluster (cyan), the number of clusters with more than one subject (magenta). (C) The clustering results of the iteration #516 with 9 subjects identified integrally. Each color stands for a cluster and cluster centers are marked with white circles.

The last example is about the high speed rail transport network in China. We hope to find out the importance of cities or whether they are transport hubs by analyzing the connectivity among all the stations. The data was downloaded from the website (ref.16) in May, 2016. Due to geological and geographical factors, the high speed railway stations are primarily distributed in the east of China. Here the similarity $s_{ij}$ is set to the number of direct routes between station $i$ and $j$ without any intermediate station. $s_{ii}$ is weighed according to following rules: If station i is the beginning or terminal station of one route, $s_{ii}$ is added by the number of stations on the route in order to reflect its importance; if station i is an intermediate station of one route, $s_{ii}$ is added by 2. Fig.4A shows the number of connection centers versus the iteration number. The curve descends sharply at the beginning, and then it begins to decrease slowly as the number of iterations increases. When k = 8, eighteen cities are chosen as connection centers (Fig.4B), which are all important cities, including provincial capitals, busy seaports and area economy centers. After some iterations, a short-lasting platform

($13 \leq k \leq 15$) emerges and the number of connection centers is reduced to 9 (Fig.4C). When $21 \leq k \leq 24$, a new locally stable result with four connection centers (Shanghai, Chengdu, Guangzhou and Beijing) is presented (Fig.4D). Interestingly, they just correspond to the four centers in the eastern, western, southern and northern part of China. In the following iterations, when $25 \leq k \leq 42$, a steady platform with a large interval appears in the curve. In this interval, the number of connection centers falls to 2 and the selected cities are, unsurprisingly, Beijing and Shanghai (Fig 4E), which are the biggest and most important cities in China. As the capital, Beijing is connected by widely distributed cities and its sphere of influence appears greater. By contrast, the number of stations assigned to Shanghai is significantly less than those assigned to Beijing. However, Shanghai finally defeats Beijing and is chosen as the ultimate connection center of all the stations. This can be attributed to the two following facts. First, Shanghai itself has a population of about 25 million, and it is the most economically developed city in China. Second, Shanghai is the center of the Yangtze River Delta which is the most prosperous and advanced region in China. The metropolitan area of the Yangtze River Delta contains some of the most economically dynamic cities such as Nanjing, Hangzhou, Suzhou, Wuxi, Ningbo, Wenzhou and Hefei, which makes the distribution of high speed railway routes in this region far more dense than those in other regions. It should be noted that one city may be assigned to a more remote center because the assignment of class in CCE is based on the actual high speed railway routes rather than the geographical adjacency of the cities. For instance, one city (Dandong) marked in blue is assigned to Beijing though it is geographically nearer to Dalian (Fig.4B), because the routes between Beijing and Dandong are twice as many as those between Dalian and Dandong.).

For many clustering problems, the true number of clusters is actually ambiguous due to the difference in the cluster scale. In general, the larger the scale is, the smaller the number of categories is, and vice versa. Most clustering algorithms work only in those cases where the scale is specified in advance and ends with an unreasonable cluster result if the scale selection is not appropriate. CCE presents a natural strategy for clustering problems and it can intelligently provide all reasonable results from local to global scale. Additionally, when the pairwise similarity matrix of the data is given, the implementation of CCE involves only the power of the matrix and does not require any manual interference, and thus is very simple and efficient. We believe that CCE actually reveals the inherent mechanism of clustering problems and will have a far-reaching impact on the region of data processing.

□

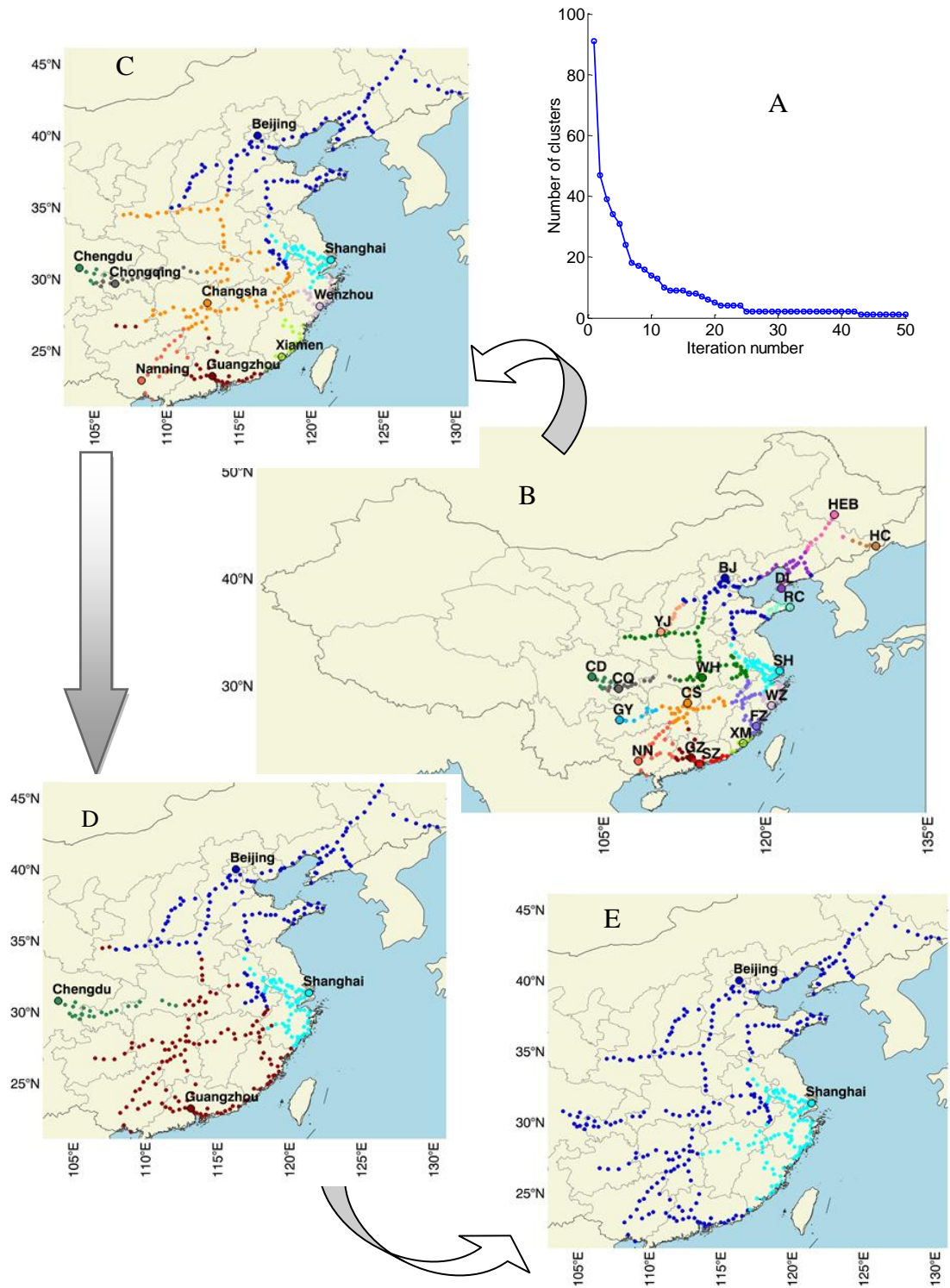

**Figure 4 | the high speed railway net in China.** CCE is used to explore the connection centers of all the stations. (A) The number of clusters versus iteration number ($k$). (B) Eighteen connection centers and the corresponding cluster result when $k = 8$. (C) Nine connection centers and the corresponding cluster result when $13 \leq k \leq 15$. (D) Four connection centers (Shanghai, Chengdu, Guangzhou and Beijing) and the corresponding cluster result when $21 \leq k \leq 24$. (E) Two connection centers (Shanghai and Beijing) and the corresponding cluster result when $25 \leq k \leq 42$.

$\mathbf{u}^0$, if it is not the eigenvector of the similarity matrix $\mathbf{S}$, there is always a large enough positive integer $k$ and a constant c, which makes the following equation hold

$$\mathbf{S}^k \mathbf{u}^0 = c\mathbf{u}_1 \tag{3}$$

Let $\mathbf{I}_n \in \mathbb{R}^{n \times n}$ be $n \times n$ identity matrix, and we choose each of its column vectors $\mathbf{i}_1, \ldots \mathbf{i}_n$ as the initial vector $\mathbf{u}^0$. Then, when $k$ approaches infinity, there must exist *n* constant, $c_1, c_2, \ldots, c_n$, making the following equations hold,

$$\begin{cases} \mathbf{S}^k \mathbf{i}_1 = c_1 \mathbf{u}_1 \\ \ldots \ldots \\ \mathbf{S}^k \mathbf{i}_n = c_n \mathbf{u}_1 \end{cases} \tag{4}$$

The equations above can be rewritten as the following matrix form:

$$\mathbf{S}^k = \mathrm{diag}(c_1, c_2, \ldots, c_n)\mathbf{u}_1 \tag{5}$$

Due to the symmetry of $\mathbf{S}^k$, the equation below can be easily derived for any $i$ and $j$:

$$\mathrm{u}_{1j}/\mathrm{u}_{1i} = \sqrt{s_{jj}^{(k)}/s_{ii}^{(k)}} \tag{6}$$

Which indicates that,

$$\mathrm{diag}(\sqrt{\mathbf{S}^k}) \propto \mathbf{u}_1 \qquad \square$$

# Supplementary Materials for

## Clustering by connection center evolution

Xiurui Geng    and    Hairong Tang

**This PDF file includes:**

    Materials and Methods

    Figures S1 to S7

    References

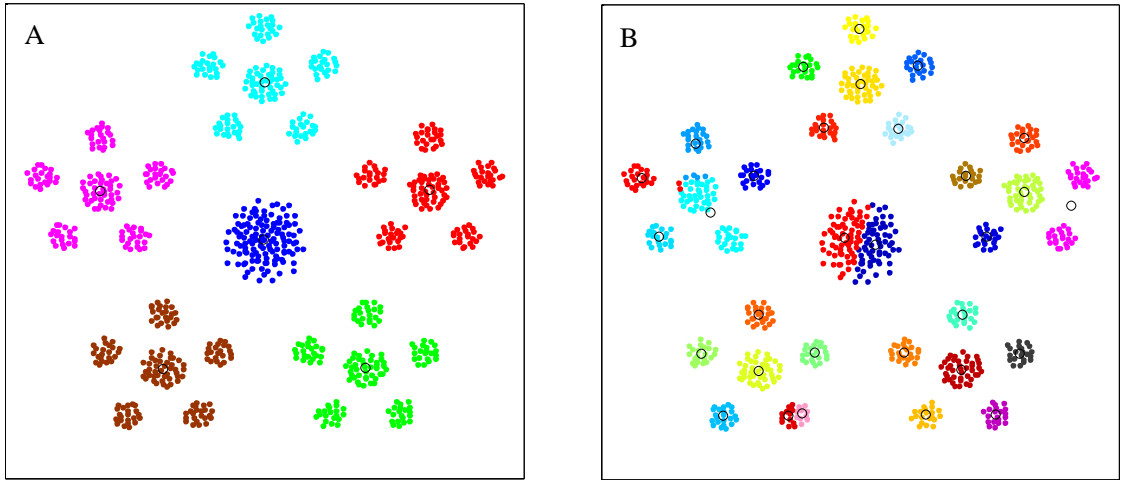

**Figure | S1**

Comparison with k-means when processing the synthetic data set as shown in Figure 2. CCE can tell us all reasonable numbers of clusters (6 and 31). For k-means, however, the number of clusters must be specified in advance. Here we suppose that the exact number of clusters is known for k-means. (A) When the number of clusters is set to 6, k-means can group the points correctly. (B) When the number of clusters is set to 31, the solution with the lowest cost value among 1000 times running is shown. Unfortunately, the result is still unreasonable.

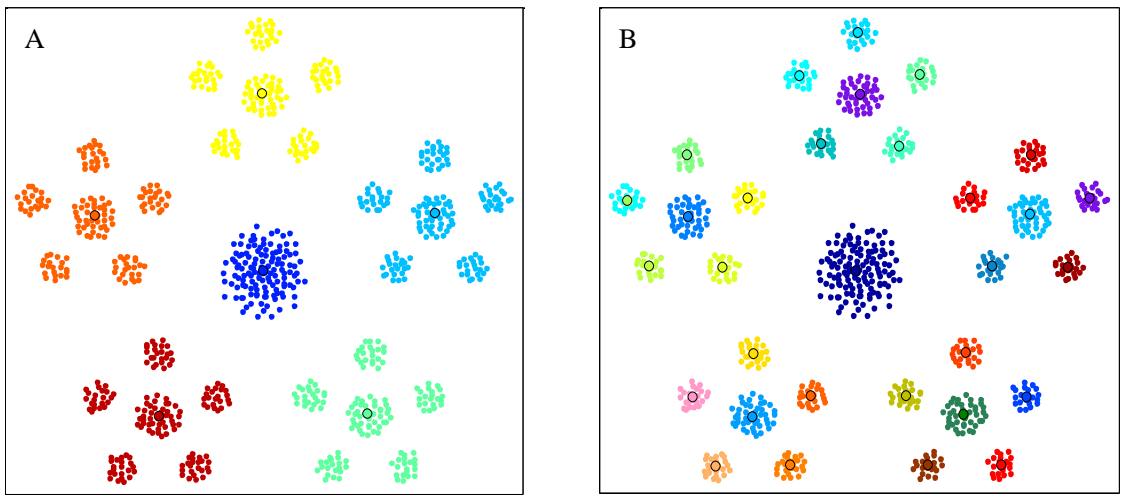

**Figure | S2**

Comparison with mean-shift when processing the synthetic data set as shown in Figure 2. CCE can intelligently provide all reasonable results from local to global scale. Mean-shift only gives one result for one specified bandwidth which corresponds to one fixed scale. In order to obtain 6 clusters and 31 clusters above, two different bandwidths must be elaborately given in advance. (A) A Gaussian kernel with twice variance of that in CCE. (B) A Gaussian kernel with the same variance as that in CCE.

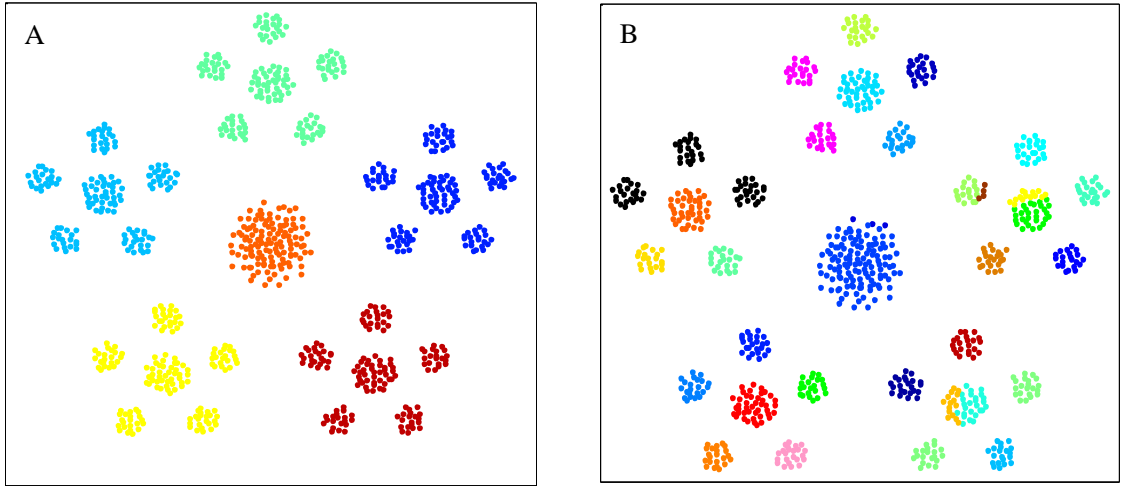

**Figure | S3**

Comparison with graph Laplacian spectral cluster method in refs.4,5 when processing the synthetic data set as shown in Figure 2. Different from CCE, this spectral method needs the number of clusters to be set in advance. Moreover, its result of 31 clusters is also not perfect. (A) The number of clusters is set to 6. (B) The number of clusters is set to 31.

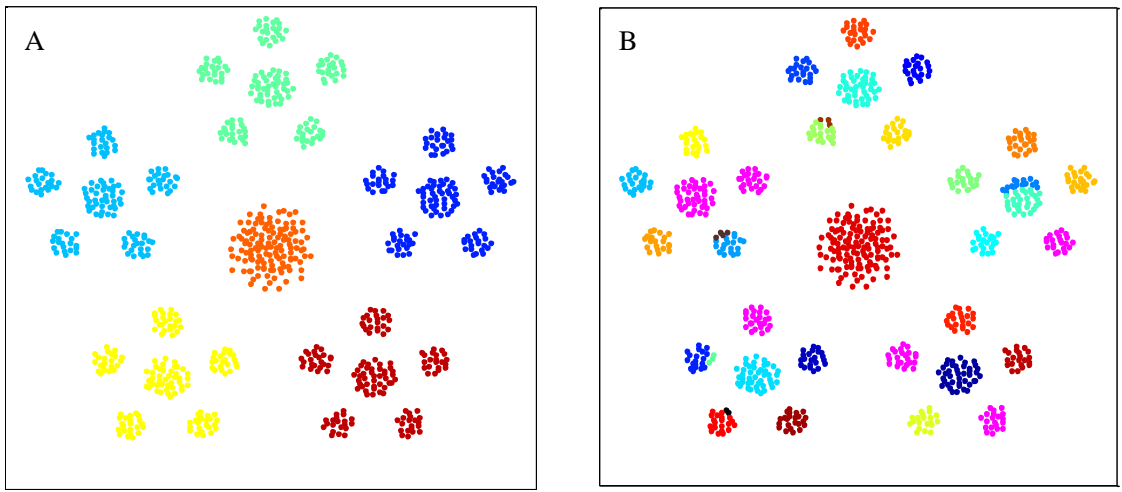

**Figure | S4**

Comparison with spectral cluster method in[6] when processing the synthetic data set as shown in Figure 2. Like graph Laplacian spectral cluster method, this spectral method also needs the number of clusters to be set in advance. Likewise, its result of 31 clusters is still not perfect. (A) The number of clusters is set to 6. (B) The number of clusters is set to 31.

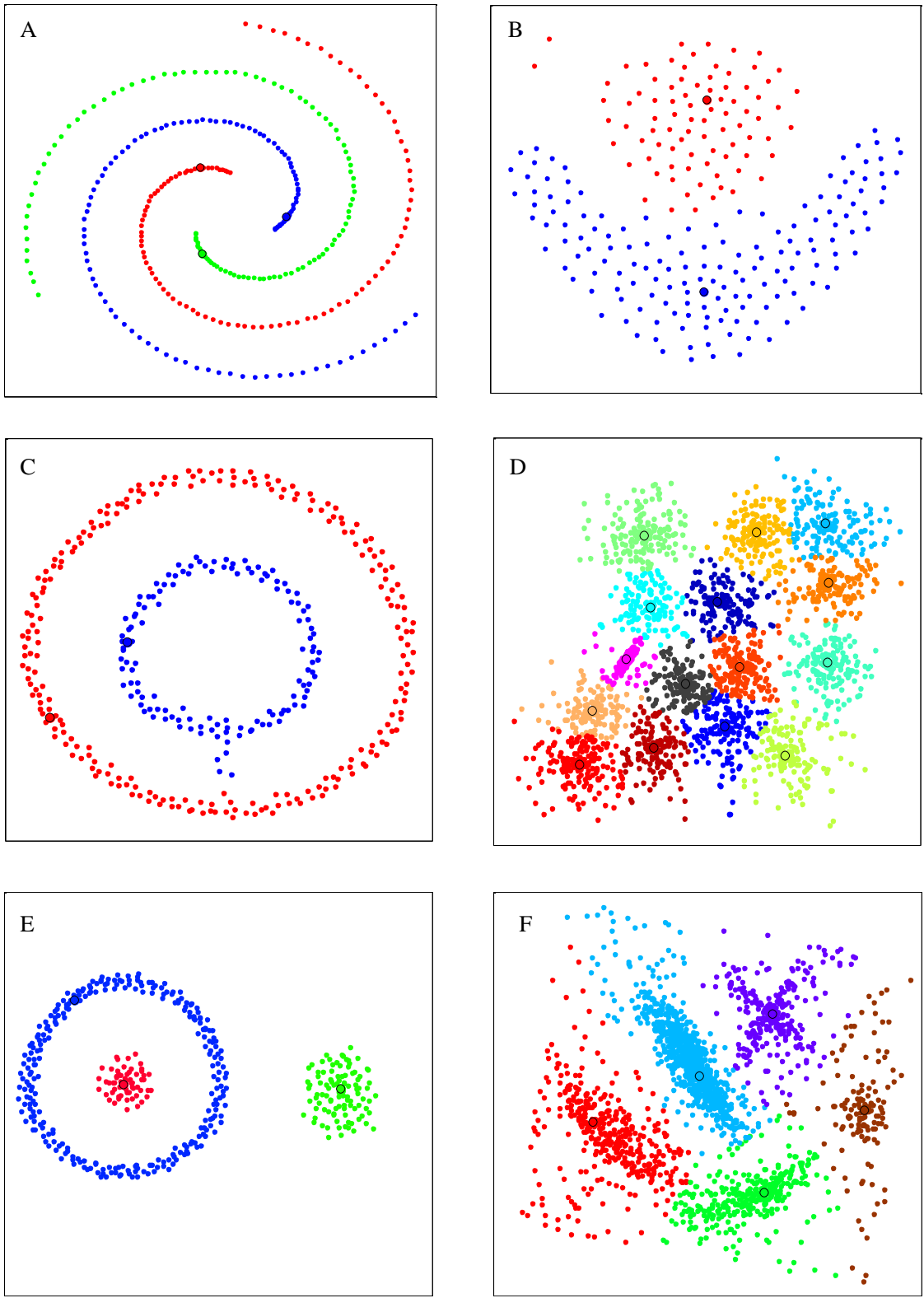

**Figure | S5**

Cluster results of CCE for some widely used synthetic data sets[1,2,3]. All the results can be indicated by the stable platforms appearing in the iteration process.

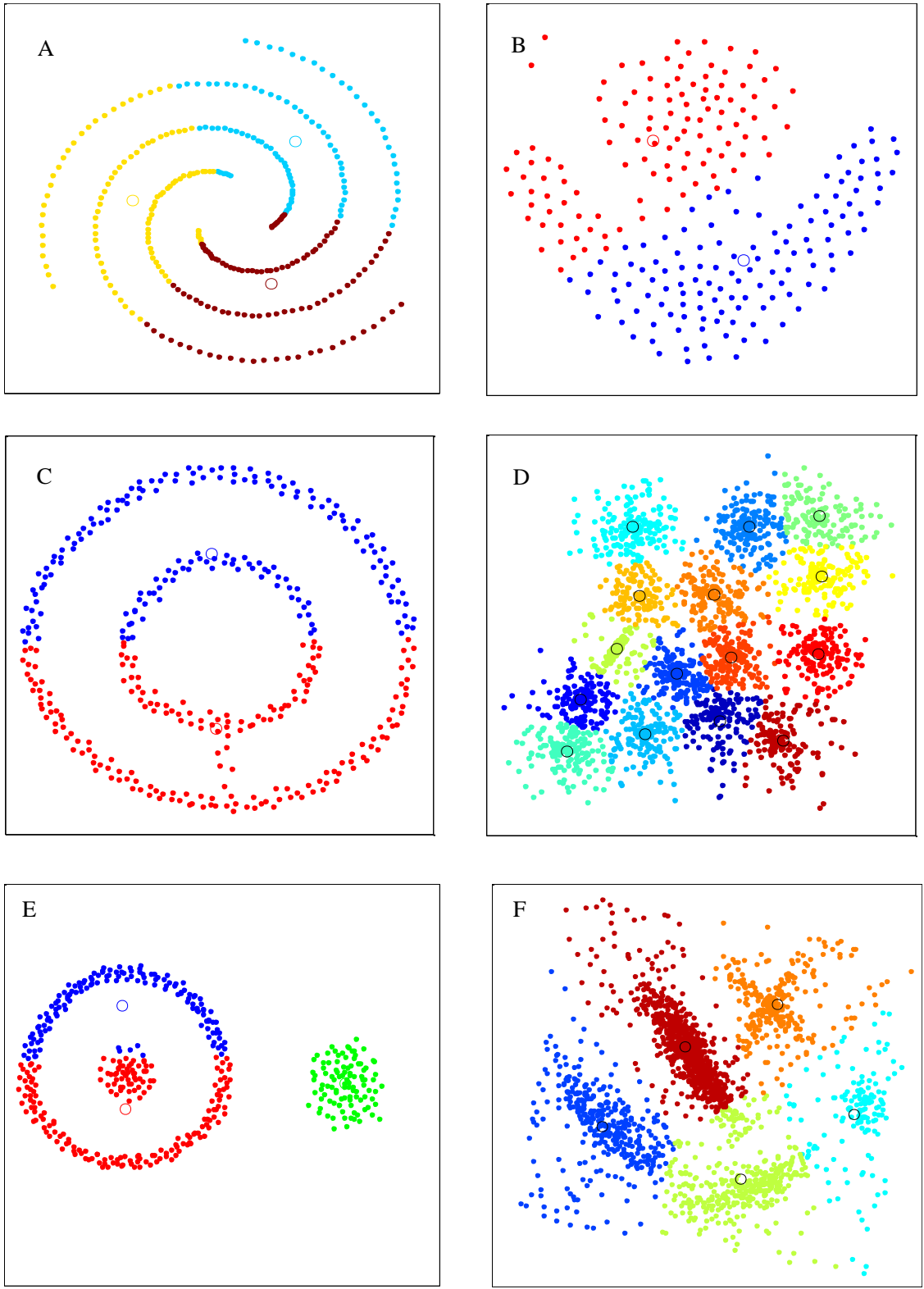

**Figure | S6**

Cluster results of K-means for above synthetic data sets.

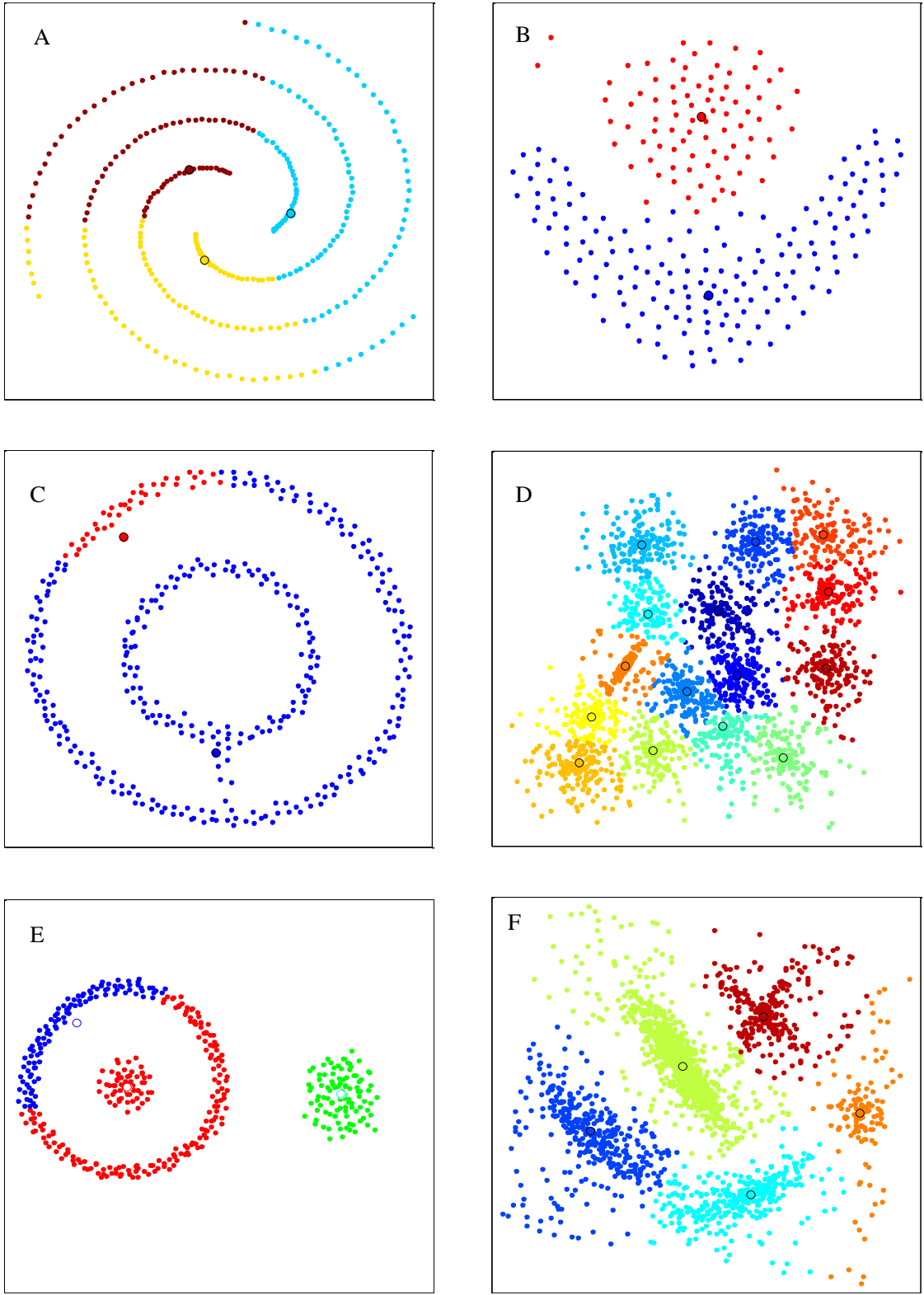

**Figure | S7**

Cluster results of mean-shift for above synthetic data sets.